\documentclass[runningheads]{llncs}

\usepackage{hyperref}
\usepackage{graphicx}
\usepackage{amsmath}
\usepackage{amssymb}
\usepackage{xspace}

\newcommand\cupEq{\protect{~\cup{\kern -0.5em}=~}}
\newcommand\ncupEq{\protect{~\#{\kern -0.3em}\cup{\kern -0.5em}=~}}
\newcommand{\ELp}{$\mathcal{E \kern -0.2em L}^+$} 
\newcommand{\ELpp}{$\mathcal{E \kern -0.2em L}^{++}$}

\newcommand{\Protege}{Prot\'{e}g\'{e}\xspace}

\newcommand{\lte}{\mathop{\leq}}

\title{OWLAx: A \Protege Plugin to Support Ontology Axiomatization through Diagramming}
\titlerunning{OWLAx: A Prot\'eg\'e Plugin to Support Ontology Axiomatization}

\author{Md. Kamruzzaman Sarker\inst{1} \and Adila A. Krisnadhi\inst{1,2} \and Pascal Hitzler\inst{1}}
\institute{Wright State University, OH, USA \and Universitas Indonesia, Depok, Indonesia}

\authorrunning{Sarker, Krisnadhi, Hitzler}

\begin{document}

\maketitle

\begin{abstract}
  Once the conceptual overview, in terms of a somewhat informal class diagram, has been designed in the course of engineering an ontology, the process of adding many of the appropriate logical axioms
  is mostly a routine task. We provide a Prot\'eg\'e\footnote{http://protege.stanford.edu/} plugin which supports this task, together with a visual user interface, based on established methods for ontology design pattern modeling.
\end{abstract}


\section{Motivation}\label{sec:motivation}

When modeling with domain experts, particularly when they do not possess intimate knowledge about ontology engineering, it is in our experience best to use a visual approach to first design a
conceptual overview of ontology modules (or corresponding content ontology design patterns), 
in the form of class diagrams~\cite{HitzlerJK15}. We have found it most effective to use non-electronic means for this, such as whiteboards and flipcharts, as they readily support a natural flow of
discussion without assuming any prior knowledge of particular software tools.

The ontology engineers in the modeling team will of course keep track of the precise meaning of each part of the diagram, so that they can convert their insights into exact specifications, i.e.,
axioms for the ontology. This conversion can then, based on the class diagram and the discussions during the modeling sessions, in our experience mostly be done by the ontology engineers without a lot
of required further interaction with the domain experts. For documentation (or publication) purposes, the class diagram will usually also be redrawn using appropriate software tools.

In our experience, based on the class diagram and the discussions with domain experts during its design, it is mostly a routine, albeit somewhat tedious, task to write down appropriate axioms for an
ontology module in an ontology editing tool. Most axioms in fact arise out of a systematic exploration of the class diagram. In order to simplify this part of the work, we have cast this systematic
exploration into a Prot\'eg\'e plugin, which we describe herein. Of course, some axioms -- arguably the more interesting and more critical ones -- will not come up as candidates during our systematic
exploration, and so will have to be added manually. Nevertheless, our plugin is helpful in making the task of adding many routine axioms much quicker and less error prone. More information about the
plugin is located at \url{http://daselab.org/content/ontology-axiomatization-support}.



\section{OWLAx: Description and Main Functionalities}\label{sec:candaxioms}



\begin{figure}[tp]
  \centering
  \includegraphics[width=\textwidth]{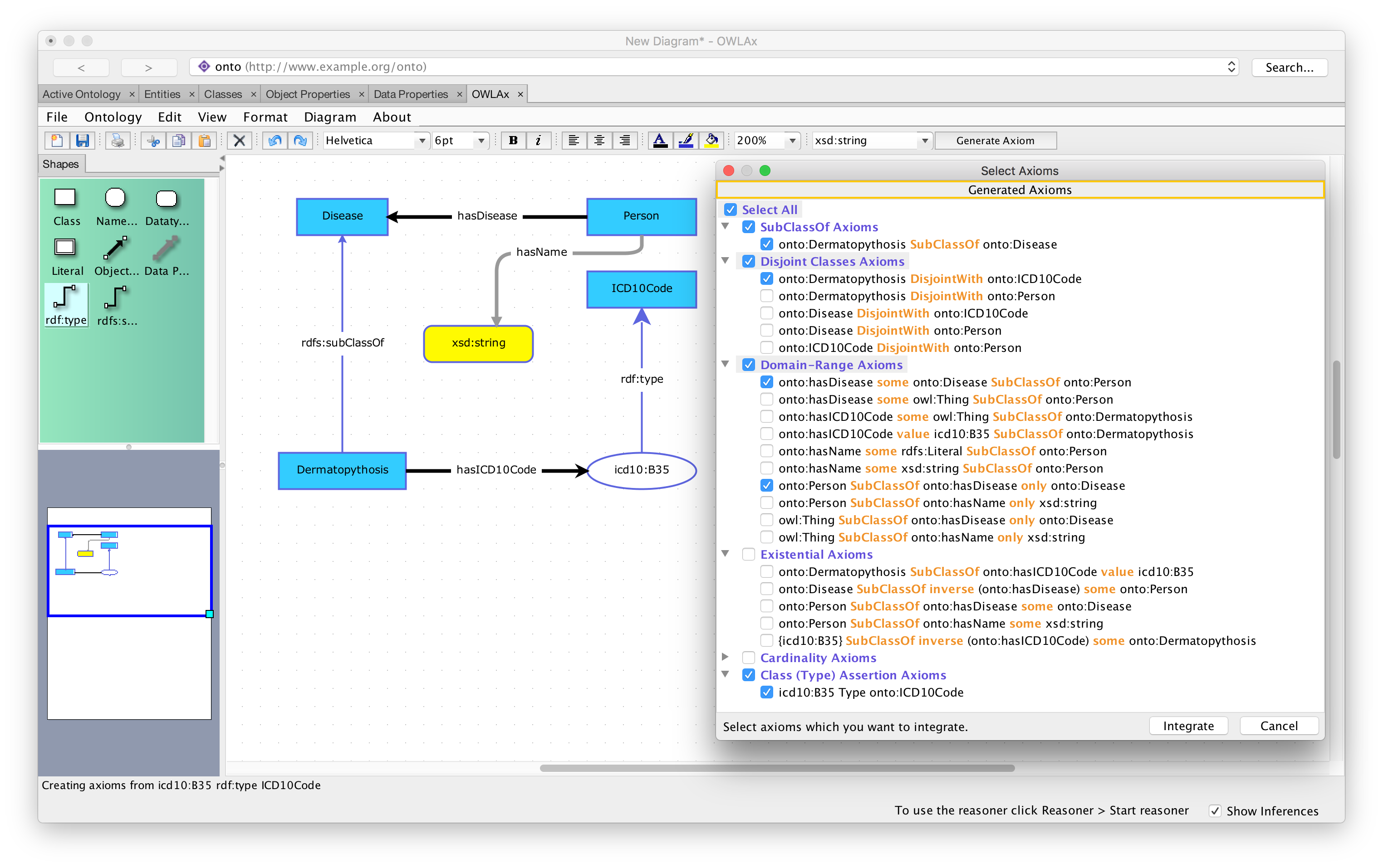}
  \caption{OWLAx UI when ``Generate Axioms'' command is executed.}
  \label{fig:plugin}
\end{figure}

The plugin provides an interface for drawing a \emph{class diagram}, and a command (accessible through an item in the menu or a button in the toolbar) to generate axioms from the given diagram, which
are to be \emph{added} into the currently active ontology. As seen in Fig.~\ref{fig:plugin}, the class diagram itself is composed of \emph{nodes} and \emph{edges}. A node in the class diagram
represents either a \emph{class}, \emph{datatype}, \emph{individual name}, or \emph{literal}. Meanwhile, an edge represents either an \emph{object property}, \emph{data property}, the \emph{typing}
relation (i.e., \textsf{rdf:type}), or the \emph{subclass} relation (i.e., \textsf{rdfs:subClassOf}). A pallette on the left side of the interface provides the user with those nodes and edges, which
can be dragged and dropped onto the canvas.

In the following, $X \xrightarrow{P} Y$ means that there is a directed edge $P$ from a node $X$ to another node $Y$ in the class diagram. Also, $A$ and $B$ denote class names, $M$ a datatype, $c$ a
named individual, $\ell$ a literal, $R$ an object property, and $Q$ a data property. The plugin enforces the class diagram to contain at least one node, and if there is an edge, it only allows the
following node-edge-node configurations: $A \xrightarrow{R} B$, $A \xrightarrow{R} c$, $A \xrightarrow{Q} M$, $A \xrightarrow{Q} \ell$, $c \xrightarrow{\textsf{rdf:type}} A$, and
$A \xrightarrow{\textsf{rdfs:subClassOf}} B$.  We do not aim to represent all possible relationships between components of the class diagram above because in our experience when modeling, the class
diagram is usually considered informal, and the aforementioned node-edge-node configurations are those typically used to describe a conceptual overview when we conduct modeling \cite{HitzlerJK15}. In
fact, we do not intend to represent all possible OWL 2 constructs in the diagram unlike, e.g., Graphol \cite{DBLP:conf/semweb/ConsoleLSS14}, Graffoo \cite{DBLP:conf/esws/FalcoGPSV14}, or Ontodia
\cite{DBLP:conf/semweb/MouromtsevPEMRG15}.

From the class diagram, a user can generate several types of candidate axioms based on the relationships depicted in the class diagram. They are only \emph{candidates} since the class diagram is
informal; each candidate axiom captures one way to read a relationship, and the actual intent should typically be inquired to the domain experts while conducting the modeling. Note that from one type
relationship, more than one actual intents need to be formalized, i.e., the candidate axioms are not mutually exclusive. On the other hand, the list of candidate axioms is not exhaustive to keep it
sufficiently simple: there are obviously axioms that will not be directly generated from the class diagram, especially if it is too complex. For such axioms, one has to simply directly input them in
\Protege.

The plugin facilitates the creation of candidate axioms through a dialog box (accessible through ``Generate Axiom'' command from the menu or toolbar) that contains a checkbox of the candidate axioms
presented in the Manchester syntax. After clicking ``Integrate'', the plugin will integrate the axioms with a check-mark to the ontology. We explain some of the candidate axioms below, though we use
mainly description logic notation \cite{BCMNP07}.

Every $c \xrightarrow{\textsf{rdf:type}} A$ leads to a \emph{class assertion} $A(c)$, and $A \xrightarrow{\textsf{rdfs:subClassOf}} B$ to a \emph{subclass axiom} $A \sqsubseteq B$. Next, for every
$A \xrightarrow{R} B$, the plugin generates several types of candidate axioms. First, it generates \emph{(unscoped) domain restriction} $\exists R.\top \sqsubseteq A$ --- equivalent to
$R$~\textsf{rdfs:domain}~$A$ --- and \emph{scoped domain restriction} $\exists R.B \sqsubseteq A$. The former would be later integrated if the domain experts involved in modeling agrees that for every
pair of instances $x,y$, if $x\ R\ y$ holds, then $x$ belongs to $A$ (regardless whether or not $y$ belongs to $B$), while the latter is chosen if the domain experts agrees that if $x\ R\ y$ holds and
$y$ is known to belong to $B$, then $x$ belongs to $A$. Such agreements will be solicited from domain experts involved in the modeling for every candidate axiom. Besides domain restrictions, the
plugin also generates \emph{scoped} and \emph{unscoped range restrictions} $A \sqsubseteq \forall R.B$, $\top \sqsubseteq \forall R.B$ --- equivalent to $R$~\textsf{rdfs:range}~$B$; several
\emph{existential axioms}, e.g., $A \sqsubseteq \exists R.B$, etc.; and several \emph{functionality restrictions}, e.g., $A \sqsubseteq (\lte 1\ R.B$), etc.

Similar types of candidate axioms are generated for every $A \xrightarrow{Q} M$, $A \xrightarrow{R} c$, and $A \xrightarrow{Q} \ell$ relationships. Finally, \emph{class disjointness axioms} are
generated as candidate axioms for every pair of different classes, unless there is a path of \textsf{rdfs:subClassOf} edges in the diagram connecting one class to the other.

\section{Implementation Information and Other Features}\label{sec:system}




The plugin is implemented on top of the OWL API provided by \Protege.  The visual components are built using mxGraph.\footnote{\url{http://jgraph.github.io/mxgraph/}} The plugin allows users to save
the diagram as XML-annotated PNG, which can then be loaded again. This plugin is \emph{not} for visualizing an ontology for which there are a number of \Protege plugins already existing, but rather,
it facilitates creating graphical class diagrams \emph{inside} \Protege and provides a way to generate axioms from it. It eliminates the need to use separate tools for creating the class diagram and
writing down the axioms. In addition, the user can customize various aspects of the class diagram, e.g., coloring, size of nodes and edges, text formatting, etc., through the provided menu or by
right-clicking the corresponding graphical components.

One could use this plugin for modeling from scratch, or starting from an already created ontology. In the latter case, the plugin will not attempt create a class diagram from the ontology, and rather,
start with an empty canvas. Nevertheless, when the user wishes to generate axioms through the plugin, existing axioms that are already in the ontology will be included as part of the list of candidate
axioms, and the user can confirm whether to keep them. Finally, we hope to continue improving this plugin, particularly to support quick modeling of modular ontologies and ontology design patterns,
and furthermore, evaluate the usability of our plugin via a comprehensive user study.



\bigskip

\noindent\emph{Acknowledgements.} This work was supported by the National Science Foundation award 1017225 \emph{III: Small: TROn -- Tractable Reasoning with Ontologies}.

\bibliographystyle{splncs03}
\bibliography{all}

\end{document}